\documentclass[journal,twoside,web]{ieeecolor}
\usepackage[table]{xcolor}
\usepackage{amsmath,amssymb,amsfonts}
\usepackage{graphicx}
\usepackage{algorithm}
\usepackage{algpseudocode}  
\usepackage{comment}
\usepackage{booktabs}
\usepackage{multirow}
\usepackage{textcomp}
\usepackage{pifont}
\usepackage{cite}
\usepackage{soul}
\usepackage{tabularx}
\usepackage{xcolor}
\usepackage{tikz}
\definecolor{orcidgreen}{HTML}{A6CE39}

\newcommand{\orcidicon}[1]{%
  \textsuperscript{%
    \href{https://orcid.org/#1}{%
      \begin{tikzpicture}[baseline=-0.1em]
        \fill[orcidgreen] (0,0) circle (1.5ex);
        \node[white, scale=0.8, font=\bfseries\sffamily] at (0,0) {iD};
      \end{tikzpicture}%
    }%
  }%
}

\makeatletter
\let\NAT@parse\undefined
\makeatother

\usepackage[colorlinks=true, linkcolor=blue, citecolor=blue, urlcolor=blue]{hyperref}

\begin{document}

\title{Preparation of Papers for IEEE {\sc Transactions on Big Data} (February 2026)} 
\author{Md. Adnanul Islam \orcidicon{0009-0006-8195-1386},
Wasimul Karim \orcidicon{0009-0008-9886-8110},
Md Mahbub Alam \orcidicon{0009-0006-2732-8710},
Subhey Sadi Rahman \orcidicon{0009-0007-6914-2874},
Md. Abdur Rahman \orcidicon{0009-0004-3097-8576},
Arefin Ittesafun Abian \orcidicon{0009-0003-4451-0838},
Mohaimenul Azam Khan Raiaan \orcidicon{0009-0006-4793-5382},
Kheng Cher Yeo \orcidicon{0000-0002-0453-3248},
Deepika Mathur \orcidicon{0000-0001-6247-7996},
Sami Azam \orcidicon{0000-0001-7572-9750}
\thanks{Manuscript submitted: 28th February, 2026.}
\thanks{(Corresponding author: Mohaimenul Azam Khan Raiaan, Sami Azam).}
\thanks{This work did not involve human subjects or animals in its research.}
\thanks{Md. Adnanul Islam, Wasimul Karim, Md. Mahbub Alam, Subhey Sadi Rahman, Md Abdur Rahman and Arefin Ittesafun Abian are affiliated with the Department of Computer Science and Engineering, United International University, Dhaka 1212, Bangladesh  and also with Applied Artificial INtelligence and Intelligent Systems (AAIINS) Laboratory, Dhaka 1217, Bangladesh (e-mail: mislam221096@bscse.uiu.ac.bd, wkarim211105@bscse.uiu.ac.bd, malam212082@bscse.uiu.ac.bd, srahman212074@bscse.uiu.ac.bd, mrahman202260@bscse.uiu.ac.bd, aabian191042@bscse.uiu.ac.bd).}
\thanks{Mohaimenul Azam Khan Raiaan is with the Department of Data Science and Artificial Intelligence, Monash University, 3800, Melbourne, Australia, and also with Applied Artificial INtelligence and Intelligent Systems (AAIINS) Laboratory, Dhaka 1217, Bangladesh (e-mail: mohaimenul.raiaan@monash.edu).}
\thanks{Kheng Cher Yeo and Sami Azam are with the Faculty of Science and Technology, Charles Darwin University, Darwin, NT 0810, Australia (e-mail: Charles.Yeo@cdu.edu.au, sami.azam@cdu.edu.au).}
\thanks{Deepika Mathur is with the Faculty of Arts and Society, Charles Darwin University, Casuarina, Darwin, NT 0810, Australia (e-mail: Deepika.Mathur@cdu.edu.au).}}


\title{Learning to Weigh Waste: A Physics-Informed Multimodal Fusion Framework and Large-Scale Dataset for Commercial and Industrial Applications}

\markboth{IEEE Transactions on Big Data, Vol. 00, No. 0, February 2026}
{Islam \MakeLowercase{\textit{et al.}}: IEEE Transactions on Big Data}

\maketitle

\begin{abstract}
Accurate weight estimation of commercial and industrial waste is important for efficient operations, yet image-based estimation remains difficult because similar-looking objects may have different densities, and the visible size changes with camera distance. Addressing this problem, we propose Multimodal Weight Predictor (MWP) framework that estimates waste weight by combining RGB images with physics-informed metadata, including object dimensions, camera distance, and camera height. We also introduce Waste-Weight-10K, a real-world dataset containing 10,421 synchronized image–metadata collected from logistics and recycling sites. The dataset covers 11 waste categories and a wide weight range from 3.5 to 3,450 kg. Our model uses a Vision Transformer for visual features and a dedicated metadata encoder for geometric and category information, combining them with Stacked Mutual Attention Fusion that allows visual and physical cues guide each other. This helps the model manage perspective effects and link objects to material properties. To ensure stable performance across the wide weight range, we train the model using Mean Squared Logarithmic Error. On the test set, the proposed method achieves 88.06 kg Mean Absolute Error (MAE), 6.39\% Mean Absolute Percentage Error (MAPE), and an $R^2$ coefficient of 0.9548. The model shows strong accuracy for light objects in the 0–100 kg range with 2.38 kg MAE and 3.1\% MAPE, maintaining reliable performance for heavy waste in the 1000–2000 kg range with 11.1\% MAPE. Finally, we incorporate a physically grounded explanation module using Shapley Additive Explanations (SHAP) and a large language model to provide clear, human-readable explanations for each prediction.
\end{abstract}


\section{Introduction}
\IEEEPARstart{A}{ccurate} estimation of commercial and industrial (C\&I) waste weight is important, as C\&I activities generate millions of tonnes of waste worldwide every year, and waste collection and transport can generate up to 70\% of the total waste management cost \cite{krishnan2021current, boskovic2016calculating}. C\&I waste refers to all waste generated by commerce and industry (including manufacturing, mining, and utilities), such as packaging, paper, metals, and catering waste, while excluding household municipal solid waste (MSW) and construction and demolition (C\&D) waste \cite{woodard2021waste, lupa2011use}. Monitoring waste through weighing data shows how much waste is produced, and provides reliable information for planning waste management, resource use, and collection services \cite{korhonen2015waste}. Direct manual sorting of waste is slow, costly, and hazardous for workers, and manual audits provide only short-term feedback that is insufficient for long-term planning, collectively highlighting the need for faster, safer, and more efficient monitoring methods \cite{wagland2012development, martinez2022vision, koskinopoulou2021robotic, chen2021looking}. While vision provides valuable information for characterizing waste, an object’s weight cannot be accurately determined from visual cues alone \cite{navarro2023visuo}. Objects with similar shapes may have very different weights due to differences in material density and irregular shapes \cite{diaz2022simultaneous}. This challenge is intensified by variations in object size in images, which depend on the camera–object distance \cite{wang2025image2mass++}. Estimating an object’s weight using only visual information is more difficult because its internal density cannot be detected from the outside, which makes appearance-based weight prediction unreliable in real-world situations \cite{mavrakis2020estimation}. Moreover, reliance on single-camera imagery limits the generalization and robustness of image-based models for waste weight estimation, due to the absence of geometric and scale information \cite{wang2025image2mass++}.

Over the years, researchers have explored vision-based methods in various data formats, including text, vision, and signal \cite{xuan2025vista, zhu2024vision+}. Methods for estimating waste objects have also evolved from early approaches in constrained settings to more recent techniques designed for complex and realistic scenarios \cite{li2021automatic}. Initial approaches were mainly tested on small and less diverse datasets, which focus on limited object types in controlled environments and struggled to handle the wide range of object sizes, materials, and conditions found in waste management scenarios \cite{standley2017image2mass, sato2024image, yu2025using}. Representative examples include image2mass \cite{standley2017image2mass}, which separately estimated density and volume from visual and geometric cues, as well as industrial and food-focused methods designed for specific object types or constrained scenarios \cite{sato2024image,lee2025scalable}. More recent works have introduced semantic and material reasoning through material prediction modules, cross-attention fusion, and pretrained vision-language models to improve mass estimation \cite{nath2024mass, wang2025image2mass++}. In parallel, multi-view and probabilistic approaches have been explored to model physical uncertainty \cite{dong2023partial, wen2025partial}, though they require controlled data capture and high computational cost. Despite these advances, most existing methods remain limited by reliance on low-weight ranges or controlled acquisition conditions, which motivates scalable multimodal solutions that explicitly integrate visual, physical, and camera-aware information.

Despite these promising developments, the existing approaches \cite{standley2017image2mass,sato2024image,wang2025image2mass++,nath2024mass} have several notable limitations. Early approaches mainly relied on single RGB images and hand-crafted geometric features \cite{standley2017image2mass}, which made them highly sensitive to changes in object scale and camera distance. Domain-specific methods have limited applicability, being constrained to specific object types such as steel cylinders \cite{sato2024image} or food items \cite{lee2025scalable} under controlled capture conditions. More recent studies introduced material-aware modeling and semantic information through additional networks, vision models, or probabilistic methods \cite{wang2025image2mass++,nath2024mass}, typically using late fusion of metadata or employing more sophisticated mechanisms. While recent methods have made progress in multimodal integration, challenges remain in generalizing across diverse object categories, material compositions, and the wide weight ranges found in real-world C\&I environments without requiring extensive domain-specific training data. Notably, prior studies have largely overlooked the explicit integration of camera geometry with physics-informed features using attention-based fusion mechanisms to address scale ambiguities caused by perspective effects, and they have also not been evaluated on the extreme weight variations that are characteristic of C\&I waste scenarios.

To address these challenges, our work introduces Multimodal Weight Predictor (MWP), a deep learning based multimodal framework for large-scale C\&I waste weight estimation. The proposed approach explicitly integrates visual understanding with structured physical metadata, including geometric dimensions. In contrast to prior methods, our proposed model is designed to resolve scale ambiguity by grounding learning in physics-informed features that reflect real-world mass relationships. To support this objective, we introduce Waste-Weight-10K, a large-scale dataset of over ten thousand image-metadata pairs collected from real C\&I environments. The dataset includes 11 C\&I waste categories: Automotive Scrap, Ferrous Metal, Cardboard, Rigid Plastic, Wood, General Trash, Industrial Gas Cylinder, Rubber, Appliance, Foam, and Battery, which frequently occur in C\&I waste streams and differ widely in size, shape, and material. These categories are important because their different sizes, shapes, and materials make it harder to estimate their weight and handle them automatically, and some of them are economically valuable or harmful to the environment. The proposed system integrates a Vision Transformer (ViT)–based visual encoder with a metadata encoder via a Mutual Attention Fusion mechanism, enabling bidirectional interaction between visual and physical cues and encouraging the model to focus on physically meaningful information rather than superficial visual patterns. This fusion strategy allows the framework to distinguish between visually similar objects with different densities and to correct perspective-induced distortions. As a result, our proposed model produces physically consistent weight predictions.

The proposed framework is evaluated on the Waste-Weight-10K dataset using a held-out test split to assess its performance across a wide range of C\&I waste weights. On the test set, the model achieves a Mean Absolute Error (MAE) of 88.06 kg, a Root Mean Square Error (RMSE) of 181.52 kg, and a Mean Absolute Percentage Error (MAPE) of 6.39\%, with an $R^2$ score of 0.9548, indicating strong overall predictive performance. Sample predictions show that the model produces consistent estimates across both medium and heavy objects, with errors remaining bounded relative to object scale. A detailed analysis by weight range further highlights the model’s behavior: it achieves high precision for lightweight waste with an MAE of 2.38 kg and a MAPE of 3.1\%, while maintaining stable relative accuracy for heavier objects despite naturally increasing absolute error. For very heavy waste of range 1000–2000 kg, the model records a MAPE of 11.1\%, demonstrating that scale-invariant optimization effectively prevents bias toward large-mass samples. While some existing methods report lower absolute errors within narrow object categories or limited weight ranges, the proposed framework achieves a strong balance between accuracy and scalability by explicitly modeling physical scale and camera geometry, enabling stable performance across diverse C\&I waste types and large variations in object mass.

The main contributions of this work can be summarized as follows:
\begin{itemize}
    \item We present a new multimodal deep learning approach for estimating the weight of C\&I waste by combining visual information with physics-informed metadata.
    \item We introduce the Waste-Weight-10K dataset, a large and diverse benchmark that includes paired images and physical measurements collected from real C\&I settings.
    \item We propose a Mutual Attention Fusion mechanism that helps the model balance visual cues and physical knowledge, allowing it to better handle object scale and material density.
    \item We show strong performance across a wide range of waste weights and provide explanations grounded in physical reasoning to support safe and interpretable use in practice.
\end{itemize}

The remainder of the paper is organized as follows. Section \ref{sec2} reviews prior work in waste management using computer vision and weight estimation techniques. Section \ref{sec:dataset} details the Waste-Weight-10K dataset acquisition protocol, statistical characteristics, and preprocessing pipeline. Section \ref{sec:methodology} describes the proposed Multimodal Weight Predictor architecture, including feature engineering, encoder designs, attention fusion mechanisms, and training strategies. Section \ref{sec:experimental_analysis} presents experimental results, including ablation studies and performance analysis across weights. Section \ref{sec6} provides an in-depth discussion of model behavior, limitations, and future directions. Finally, Section \ref{sec7} concludes the paper.

\section{Related Work}
\label{sec2}
In this section, we review recent advances in weight estimation from visual data and highlight the evolution from single-modal vision-based approaches to multimodal frameworks. We examine category-specific methods, material-aware architectures, and attention-based fusion mechanisms, as these approaches have been proposed to handle scale ambiguity and improve generalization in physical property estimation.

\subsection{Vision-Based Mass Estimation}

Early works on vision-based mass estimation mainly focused on controlled settings, where object categories were limited and hand-crafted features were used to infer weight from visual information \cite{standley2017image2mass,sato2024image,lee2025scalable}. Standley et al. \cite{standley2017image2mass} introduced the image2mass framework, which modeled mass as the product of density and volume using a two-tower architecture. A geometry module predicted low-resolution thickness masks from synthetic ShapeNet data and extracted 14 geometric features such as object dimensions, maximum thickness, and occupancy ratios. Evaluated on approximately 150K Amazon products, the method achieved an mALDE of 0.470 and an mAPE of 0.651 on a 924-item test set.  Following this work, several methods targeted narrow industrial scenarios. Sato et al. \cite{sato2024image} proposed a steel cylinder weight estimation system based on traditional image processing techniques rather than deep learning. It reconstructed 3D shape from eight images captured at fixed angles, extracted contours using simple thresholding, and estimated volume through cross-section integration. The final weight was computed by combining this estimate with a basic cylinder model assuming a fixed iron density. This hybrid approach achieved 91-92\% accuracy within a $\pm$10g tolerance and was computationally efficient. 

On the other hand, Lee et al. \cite{lee2025scalable} proposed a vision-based food weight estimation method that combined image segmentation with geometric feature extraction. Their approach relied on RGB images captured in a controlled setting and utilized hand-crafted 2D shape features, such as projected area and contour descriptors, to infer food volume and weight through regression models. Despite these advances, these approaches depended on manual size measurements, used synthetic thickness masks that performed poorly on thin or blank real objects, required full image coverage, worked only for cylindrical steel parts, and supported a very narrow weight range, which limited their ability to generalize.

\subsection{Material-Aware and Semantic Enhancement}

Since object mass depends on both geometry and material composition, several recent works have introduced explicit material reasoning to improve mass estimation accuracy \cite{wang2025image2mass++,nath2024mass}. Wang et al. \cite{wang2025image2mass++} extended the original image2mass framework by adding material and semantic understanding. The proposed Image2Mass++ model has four parts: a thickness module using Xception trained on synthetic ShapeNet data, a material module trained on the EMMA dataset for 182 material classes, a classification module that combines object class and size using a frozen CLIP model, and a geometry module that extracts 14 shape features. These features are combined using cross-attention, while density and volume are estimated separately and merged to predict mass. However, the method strongly depends on manually provided object dimensions. For example, when dimensions were removed, performance dropped sharply. In addition, the method could not handle unknown camera distances. Nath et al. \cite{nath2024mass} explored a lighter-weight alternative by adding material embeddings to the image2mass model. They extracted 1024-dimensional features from a pre-trained image-to-material network and fused them with mass prediction; however, the use of coarse MINC material labels, such as metal or plastic, failed to capture differences within the same material. Errors in material prediction directly affected mass estimates, and the model did not provide any measure of confidence or uncertainty. Research has been conducted on efficient data loading and metadata management techniques for large-scale dataset processing \cite{wang2025lafa, mohan2021temporal}.

\begin{table*}[ht!]
\centering

\caption{Comparative analysis of image-based mass and weight estimation methods, highlighting modality, architecture, geometric and material reasoning, scale handling, optimization strategy, and applicable weight range.}
\scriptsize
\begin{tabular}{
>{\centering\arraybackslash}p{0.9cm}
>{\centering\arraybackslash}p{1.5cm}
>{\centering\arraybackslash}p{2.8cm}
>{\centering\arraybackslash}p{2.3cm}
>{\centering\arraybackslash}p{2.5cm}
>{\centering\arraybackslash}p{2.2cm}
>{\centering\arraybackslash}p{1.8cm}
>{\centering\arraybackslash}p{1.6cm}
}
\midrule
\textbf{Study} &
\textbf{Input Modality} &
\textbf{Architecture} &
\textbf{Dataset(s)} &
\textbf{Geometry Modeling} &
\textbf{Material Modeling} &
\textbf{Scale Ambiguity Handling} &
\textbf{Weight Range} \\
\midrule

\cite{standley2017image2mass} &
Single RGB &
Two-tower CNN (density + volume) &
Amazon (150k), Household &
Explicit (thickness masks + geometry) &
Implicit (appearance-based) &
None (requires dimensions) &
Approximate  0.03-2 kg \\

\cite{sato2024image} &
Multi-view RGB &
3D reconstruction + cylinder model + regression &
100 industrial steel parts &
Explicit (full 3D reconstruction) &
Fixed (iron density) &
Multi-view only &
Not Specified \\

\cite{lee2025scalable} &
RGB &
YOLOv8 + Support Vector regression &
food ingredients images &
Calibrated 2D shape-based geometry &
Implicit (learned via regression) &
None &
Not Specified \\

\cite{nath2024mass} &
Single RGB &
Image2Mass + material embeddings &
Amazon, MINC &
Explicit (thickness + dimensions) &
Broad categorical (MINC) &
None (requires dimensions) &
0.6-6 kg \\

\cite{wang2025image2mass++} &
Single RGB &
Xception + CLIP + cross-attention &
Amazon, ABO-500 &
Explicit (thickness + geometry features) &
Explicit (182 materials) &
None (fails without dimensions) &
Not Specified \\

\textbf{Ours} &
\textbf{Single RGB + Physical \& Camera Geometry} &
\textbf{ViT + Metadata Encoder + Mutual Attention Fusion} &
\textbf{Waste-Weight-10K (C\&I)} &
\textbf{Explicit (physics-informed geometric features)} &
\textbf{Explicit (category-aware density reasoning)} &
\textbf{Explicit (distance-aware, perspective-corrected)} &
\textbf{3.5-3,450 kg} \\
\midrule
\end{tabular}
\label{tab:weight_estimation_comparison}
\end{table*}

\begin{figure*}[!ht]
    \centering
    \includegraphics[scale=0.45]{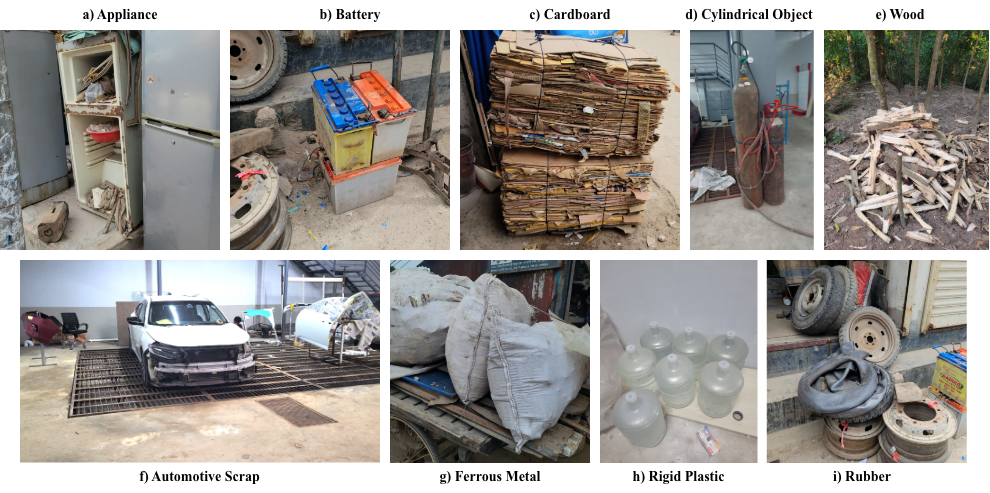}
    \caption{Representative samples from the Waste-Weight-10K dataset. The figure illustrates the diversity in object shapes, textures, and lighting conditions captured across different commercial \& industrial environments.}
    \label{fig:data_sample}
\end{figure*}

Contrary to the existing methods that focus on narrow weight ranges, such as food items \cite{lee2025scalable}, or curated online products with limited variation \cite{standley2017image2mass,wang2025image2mass++}, our framework is designed for C\&I settings and supports a much wider mass range from 3.5 to 3,450 kg. To handle this large variation, we adopt a scale-invariant Mean Squared Logarithmic Error (MSLE) loss that treats relative errors equally across all weight levels, preventing the training process from being dominated by very heavy objects, which is a common issue with standard MSE-based methods \cite{nath2024mass}. In contrast to approaches that rely on manually measured object dimensions \cite{standley2017image2mass,wang2025image2mass++,nath2024mass} and fail when such inputs are unavailable, our method incorporates camera distance and mounting height to correct perspective effects, allowing the model to better distinguish between small nearby objects and large distant ones. While multi-view techniques \cite{sato2024image} resolve scale through full 3D reconstruction, they require controlled data capture and high computational cost, making them unsuitable for real-world logistics environments. Importantly, none of the reviewed studies evaluate their methods on real-world C\&I waste datasets that cover wide material types and very large weight ranges. Our work addresses this gap by operating directly in such settings and achieving reliable performance across a 1000$\times$ weight span. Our proposed approach addresses the key limitations observed in prior weight estimation studies, as summarized in Table \ref{tab:weight_estimation_comparison}. 

\section{Dataset}
\label{sec:dataset}

To address the paucity of multimodal benchmarks in C\&I waste quantification, we introduce \textit{Waste-Weight-10K}, a heterogeneous dataset synchronizing visual perception with physical metrology. Comprising 10,421 high-resolution image-metadata pairs, this collection is specifically curated to capture the complex interplay between volumetric data and material density. By providing a holistic snapshot of C\&I payloads, the dataset directly resolves the scale ambiguities inherent in single-view weight regression.

\subsection{Composition and Acquisition}
To ensure the model works in real-world conditions, we collected this data at various logistics centers and recycling facilities instead of using a controlled laboratory. The collection process was carried out by the three authors of this study. To ensure acquisition variability, we utilized multiple mobile imaging devices to capture high-resolution images of the C\&I waste (Figure \ref{fig:data_sample}). 

\begin{figure}[!ht]
    \centering
    \includegraphics[scale=0.2]{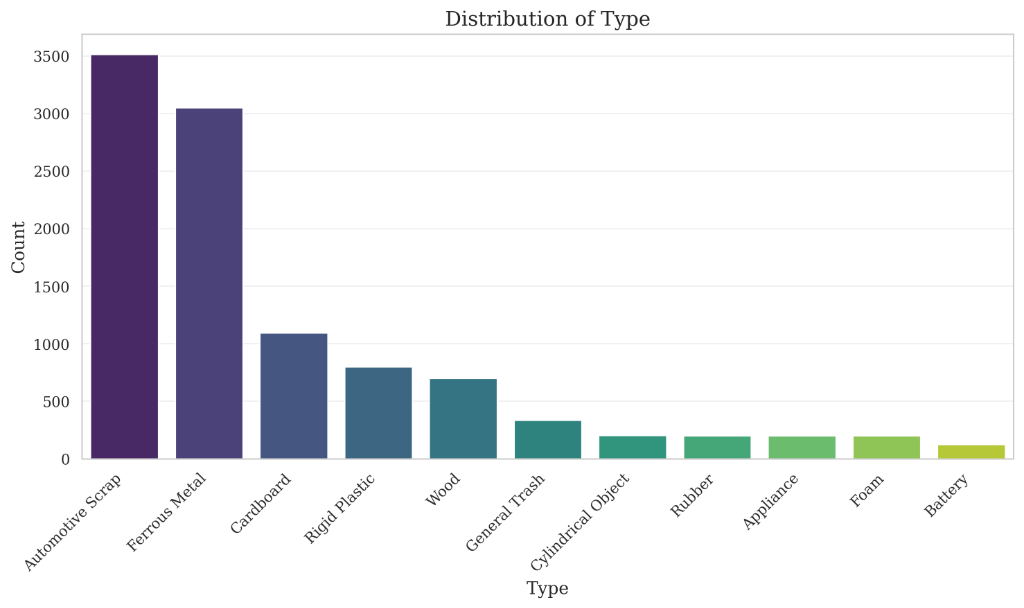}
    \caption{Distribution of Waste Categories. The dataset includes many types of metal scrap collected from C\&I environments.}
    \label{fig:type_dist}
\end{figure}

To ensure accuracy in our physical measurements, we used a professional steel tape to measure the horizontal distance from the camera, the camera's mounting height, and the three-dimensional dimensions of the waste object. After collection, we cleaned the data by standardizing category names and removing duplicates, resulting in the 11-category taxonomy shown in Figure \ref{fig:type_dist}.

\subsection{Mass Distribution and Physical Properties}
\begin{figure}[!ht]
    \centering
    \includegraphics[scale=0.24]{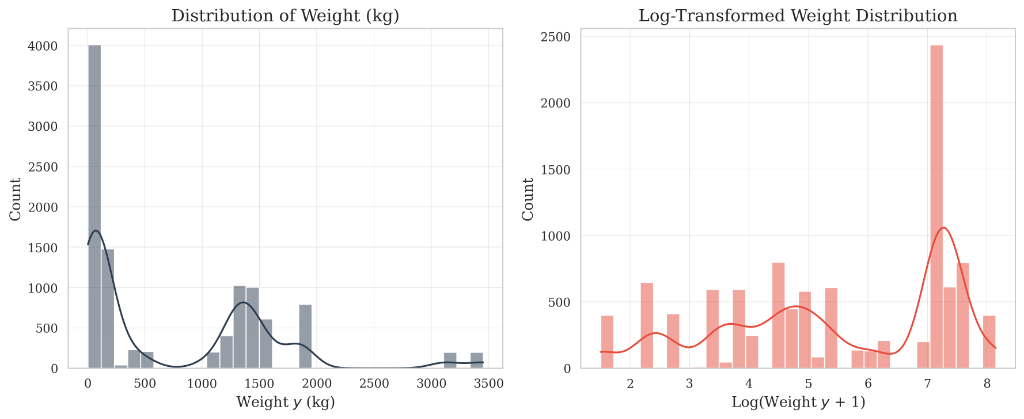}
    \caption{Weight Distribution of the Dataset. The chart on the left shows the raw weights, while the chart on the right shows the weights after a log transformation, which simplifies the training process for the model.}
    \label{fig:weight_dist}
\end{figure}

The physical weight ($y$) of objects in the dataset ranges from 3.5 kg to 3,450.0 kg, as summarized in Table \ref{tab:dataset_stats}. The average (mean) weight is 751.93 kg, while the median is 210.0 kg. Roughly 73.9\% of our samples (7,708) weigh 50 kg or more. This wide variation indicates that the dataset encompasses a significant dynamic range. Figure \ref{fig:weight_dist} shows the weight distribution of our dataset.

\begin{table}[!ht]
\centering
\small
\caption{Summary of physical and geometric measurements ($N=10,421$). Here, the geometric measurements for each object's 3D size are $L_x, L_y, L_z$, and $D_x$ and $D_y$ are the horizontal camera distance and lens height, respectively.}
\label{tab:dataset_stats}
\scriptsize
\begin{tabular}{lccccc}
\midrule
\textbf{Feature} & \textbf{Mean} & \textbf{Std} & \textbf{Min} & \textbf{Median} & \textbf{Max} \\
\midrule
Weight (kg) & 751.93 & 841.50 & 3.5 & 210.0 & 3450.0 \\
$L_x$ (m) & 96.26 & 54.26 & 12.0 & 92.0 & 180.0 \\
$L_y$ (m) & 51.27 & 26.25 & 12.0 & 47.0 & 891.0 \\
$L_z$ (m) & 45.32 & 17.10 & 6.0 & 48.0 & 85.0 \\
$D_x$ (m) & 86.55 & 35.56 & 24.0 & 85.0 & 187.0 \\
$D_y$ (m) & 55.64 & 13.43 & 12.0 & 52.0 & 126.0 \\
\midrule
\end{tabular}   
\end{table}

\subsection{Geometric and Spatial Attributes}
\begin{figure}[ht!]
    \centering
    \includegraphics[scale=0.32]{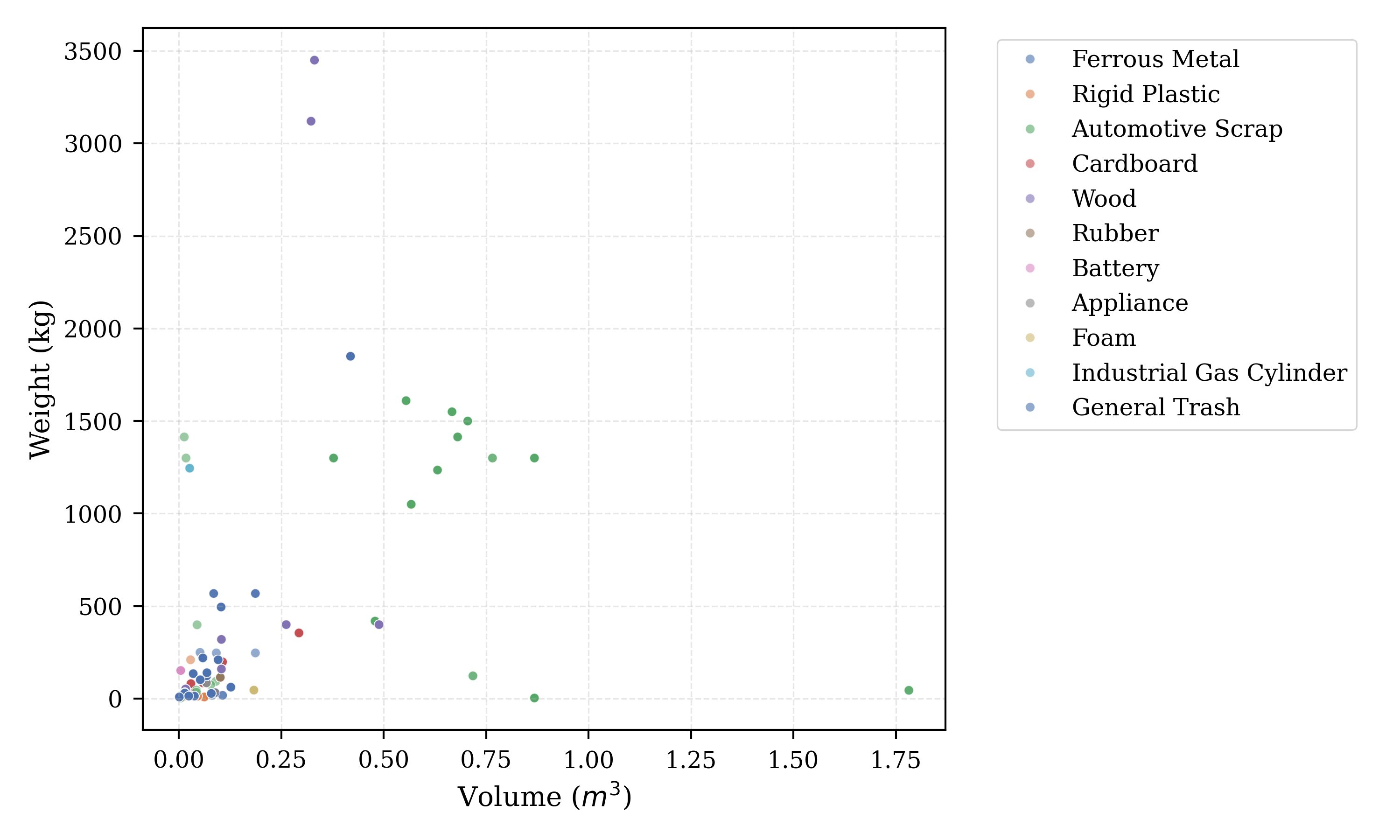}
    \caption{Scatter plot of Volume ($m^3$) versus Weight (kg). The distribution highlights the complexity of the problem: small dense objects can be heavier than large voluminous ones, necessitating a multimodal approach.}
    \label{fig:vol_vs_weight}
\end{figure}

We provide several geometric measurements for each object: its three-dimensional size ($L_x, L_y, L_z$), the horizontal camera distance ($D_x$), and the lens height ($D_y$). These values help the model understand physical scale. For instance, $D_x$ and $D_y$ allow the network to distinguish between a small object in the foreground and a large container that appears small due to perspective. As illustrated in Figure \ref{fig:vol_vs_weight}, the relationship between volume and weight is non-linear and highly category dependent, underscoring the challenge of estimating mass from visual data alone.

\subsection{Semantic Diversity}
Our dataset covers a wide spectrum of C\&I materials. As shown in Figure \ref{fig:type_dist}, the distribution is dominated by \textit{Automotive Scrap} 33.7\% and \textit{Ferrous Metal} 29.3\%. However, the dataset also includes lighter materials such as \textit{Cardboard} 10.5\% and \textit{Foam} 1.9\%. This diversity tests the model's ability to isolate material density from geometric volume. Table \ref{tab:taxonomy} details the share of each category in the cleansed taxonomy, along with their corresponding weight and volume statistics.

\begin{table}[ht!]
\centering
\scriptsize
\setlength{\tabcolsep}{5.2pt}
\caption{Final Categorization of C\&I Waste Categories with Weight Ranges.}
\label{tab:taxonomy}
\begin{tabular}{
>{\raggedright\arraybackslash}p{2.4cm}
>{\centering\arraybackslash}p{0.5cm}
>{\centering\arraybackslash}p{0.5cm}
>{\centering\arraybackslash}p{0.5cm}
>{\centering\arraybackslash}p{0.5cm}
>{\centering\arraybackslash}p{0.7cm}
>{\centering\arraybackslash}p{0.7cm}
}
\toprule
\textbf{Category} & \textbf{Count} & \textbf{Share (\%)} & \textbf{Min (kg)} & \textbf{Max (kg)} & \textbf{Min Vol ($m^3$)} & \textbf{Max Vol ($m^3$)} \\
\midrule
Automotive Scrap & 3,514 & 33.7\% & 3.5 & 1,610 & 0.0039 & 1.7820 \\
Ferrous Metal & 3,050 & 29.3\% & 9 & 1,850 & 0.0010 & 0.4189 \\
Cardboard & 1,094 & 10.5\% & 9.3 & 355 & 0.0084 & 0.2930 \\
Rigid Plastic & 799 & 7.7\% & 9 & 210 & 0.0284 & 0.0884 \\
Wood & 701 & 6.7\% & 52 & 3,450 & 0.0162 & 0.4885 \\
General Trash & 336 & 3.2\% & 14 & 28 & 0.0240 & 0.0790 \\
Industrial Gas Cylinder & 202 & 1.9\% & 1,245 & 1,245 & 0.0262 & 0.0262 \\
Rubber & 200 & 1.9\% & 115 & 115 & 0.1010 & 0.1010 \\
Appliance & 200 & 1.9\% & 85 & 85 & 0.0671 & 0.0671 \\
Foam & 200 & 1.9\% & 46 & 46 & 0.1830 & 0.1830 \\
Battery & 125 & 1.2\% & 152 & 152 & 0.0043 & 0.0043 \\
\bottomrule
\end{tabular}
\end{table}


\section{Methodology}
\label{sec:methodology}

This study proposes a multimodal deep learning framework for waste weight estimation. Our approach integrates visual features with structured metadata (such as material type and size) using a Mutual Attention Fusion module \cite{tong2024enhanced}. This design handles input heterogeneity and the large dynamic range of C\&I loads. 

\begin{figure*}[!ht]
    \centering
    \includegraphics[scale=0.06]{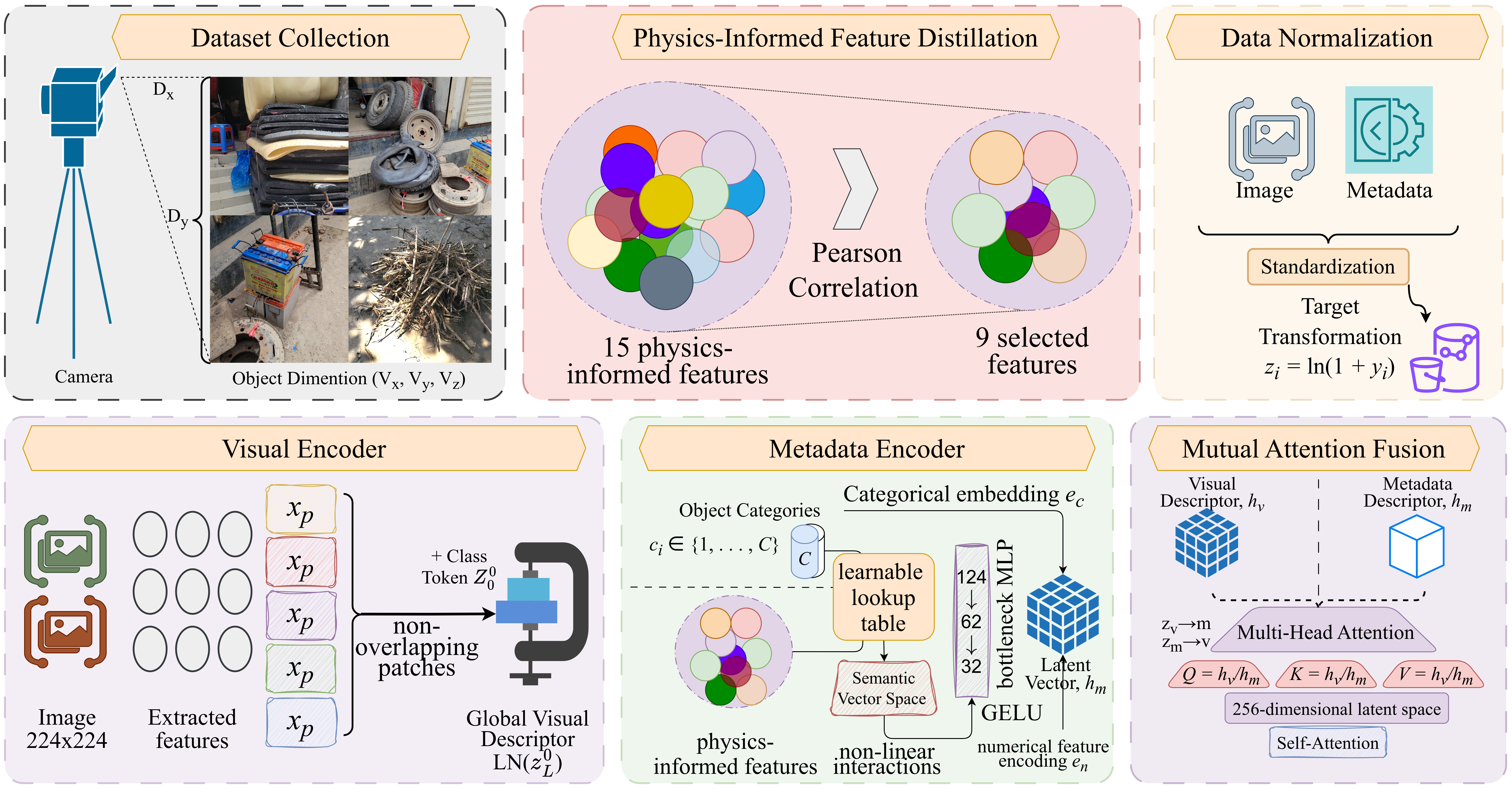}
    \caption{Overview of the proposed Multimodal Weight Predictor (MWP) framework. It integrates visual features with metadata through a Mutual Attention Fusion mechanism.}
    \label{fig:overview}
\end{figure*}

\subsection{Problem Formulation}
We treat this weight estimation task as a regression problem using a dataset $\mathcal{D} = \{(X_i, y_i)\}_{i=1}^N$. Each sample $X_i$ has two parts: the visual data $x^{(v)}_i$ (the RGB image) and the metadata $x^{(m)}_i$ (physical details such as size and camera distance). The target $y_i$ is the actual weight of the object.

Our goal is to learn a function $\mathcal{F}_\theta$ that predicts the weight based on these inputs. We want the predicted weight $\hat{y}_i = \mathcal{F}_\theta(x^{(v)}_i, x^{(m)}_i)$ to be as close as possible to the real weight $y_i$. Since the weights vary a lot (from light to very heavy), we use a special loss function called $\mathcal{L}_{MSLE}$. This helps the model predict accurately for both small and large objects. Figure \ref{fig:overview} overviews the proposed framework.

\subsection{Feature Engineering}
\label{subsec:preprocessing}

To ensure numerical stability and improve model generalization, we apply a rigorous preprocessing pipeline to both metadata and visual inputs. 

\subsubsection{Physics Informed Feature Distillation}
To bridge the gap between monocular vision and physical mass, we implemented a feature engineering pipeline grounded in classical mechanics. Weight is fundamentally a product of mass, which for C\&I objects is dictated by the relationship between volumetric geometry and material density. We derived a comprehensive set of geometric and physical descriptors from visual textures and object dimensions.

To prevent the curse of dimensionality \cite{jiang2023augmented, wei2021unsupervised}, we employed a hybrid selection inference strategy. While volumetric predictors were filtered based on strong Pearson correlation ($r$) to maximize linear predictivity, shape and perspective descriptors were retained based on physical domain principles, specifically their utility as density proxies and scale-correction factors, regardless of their global linear correlation. This physics-first approach provided 9 important indicators (Table \ref{tab:feature_engineering}).

\begin{table*}[!ht]
\centering
\small
\caption{Taxonomy of Engineered Physics-Informed Features. Selection utilizes a hybrid strategy: \textbf{Correlation ($r$)} for size magnitude and \textbf{Domain Knowledge} for material density and perspective correction.}
\label{tab:feature_engineering}
\scriptsize
\begin{tabular}{l l l l}

\midrule
\textbf{Aspect} & \textbf{Feature} & \textbf{Formula}  & \textbf{Selection Rationale (Hybrid Strategy)} \\
\midrule
\multirow{4}{*}{\textbf{Size}} & \textbf{log(Volume)} & $\ln(1 + L_x L_y L_z)$ & \textbf{Primary Signal}: High correlation ($r=0.66$). Anchor feature. \\
 & log(Surf. Area) & $\ln(1 + 2\sum L_i L_j)$ & \textit{Rejected}: Collinear with Volume ($r>0.99$). Redundant. \\
 & \textbf{log(Max Dim)} & $\ln(1 + \max(L_i))$ & \textbf{Shape Outlier}: $r=0.59$. Captures long beams/pipes distinct from cubes. \\
 & log(Geo Mean) & $\ln(1 + \sqrt[3]{L_x L_y L_z})$ & \textit{Rejected}: Mathematical duplicate of Volume ($r=1.0$). \\
\midrule
\multirow{2}{*}{\textbf{Material}} & \textbf{Compactness} & $L_{\min} / L_{\max}$ & \textbf{Density Proxy}: $r=-0.29$. Distinguishes dense bales from loose scrap. \\
 & \textbf{log(Vol/Surf)} & $\ln(1 + V/A_s)$ & \textbf{Complexity}: $r=0.64$. Measures object "hollowness". \\
\midrule
\multirow{5}{*}{\textbf{Geometry}} & \textbf{Elongation} & $L_{\max} / L_{\text{mid}}$ & \textbf{Material Indicator}: $r=0.12$. Identifying timber/pipes vs. bulk trash. \\
 & \textbf{Asp. Ratio XY} & $L_x / L_y$ & \textbf{Profile Match}: $r=-0.09$. Validates visual texture via width profile. \\
 & \textbf{Asp. Ratio YZ} & $L_y / L_z$ & \textbf{Profile Match}: $r=-0.01$. Validates visual texture via depth profile. \\
 & Sphericity & $\pi^{1/3}(6V)^{2/3} / A_s$ & \textit{Rejected}: High redundancy ($r=0.86$) with Compactness. \\
 & Flatness & $L_{\min} / L_{\text{mid}}$ & \textit{Rejected}: $r=-0.06$. Low signal-to-noise for crumpled C\&I waste. \\
\midrule
\multirow{4}{*}{\textbf{Context}} & \textbf{Surf-Sphere} & $\ln(\text{Surf}) \times \Psi$ & \textbf{Interaction}: $r=0.42$. Combines size magnitude with shape complexity. \\
 & Vol-Compact & $\ln(\text{Vol}) \times C$ & \textit{Rejected}: Redundant ($r=0.97$) with Compactness. \\
 & \textbf{log(Dist)} & $\ln(1 + \sqrt{D_x^2+D_y^2})$ & \textbf{Scale Invariance}: $r=0.36$. Critical correction for perspective distortion. \\
 & log(App. Vol) & $\ln(1 + V/D_x^2)$ & \textit{Rejected}: Implicitly learned by $Visual \times Dist$ interaction. \\
\midrule
\end{tabular}%
\end{table*}

\begin{figure*}[!ht]
    \centering
    \includegraphics[scale=0.1]{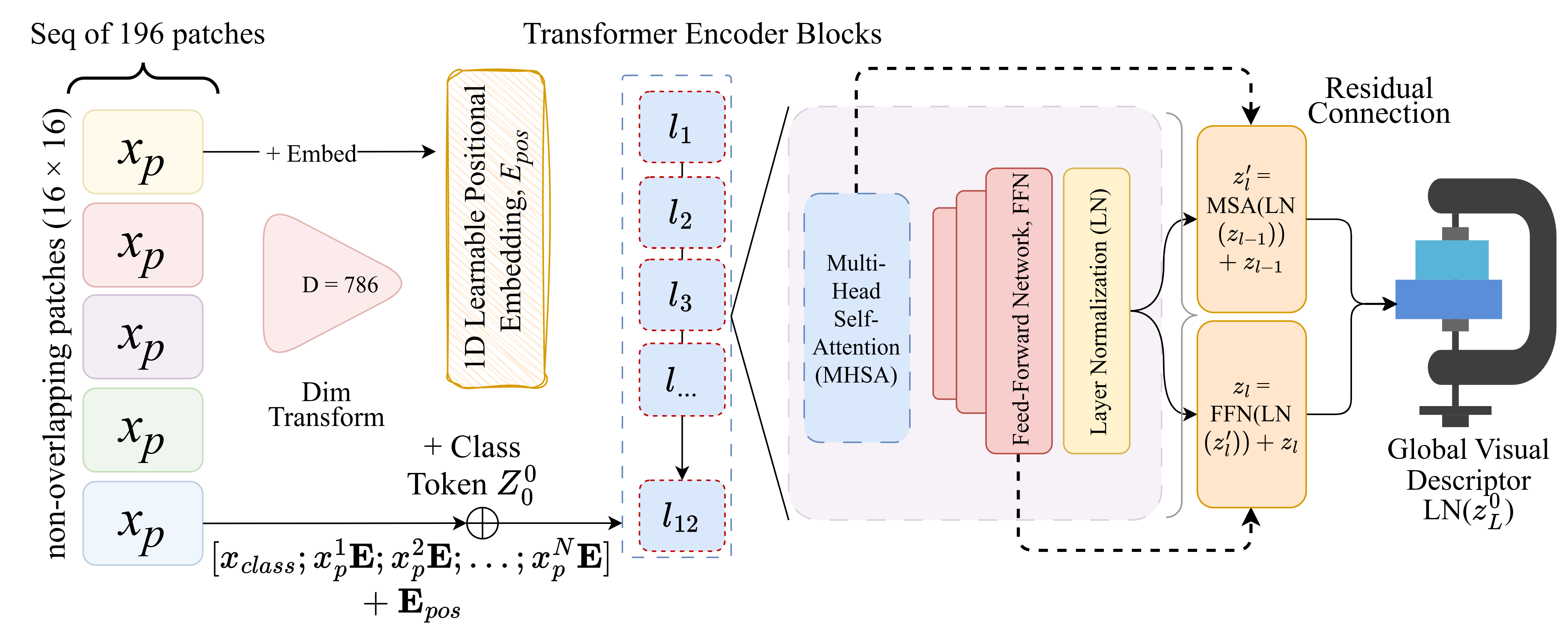}
    \caption{Architecture of our proposed extended visual transformer.}
    \label{fig:vis_enc}
\end{figure*}

The selection rationale prioritizes orthogonal physical insights. For instance, while logarithmic surface area was rejected due to extreme collinearity with volume ($r > 0.99$), "low-correlation" features such as Compactness were retained as critical classifiers for material state (distinguishing dense metallic bales from loose scrap). Crucially, the logarithmic distance feature is preserved despite negligible weight correlation, as it provides the necessary scale invariance for the visual encoder, allowing the network to distinguish between true small-scale objects and large containers appearing small due to perspective.

\subsubsection{Metadata Representation and Normalization}
Before fusion, raw metadata undergoes categorical and numerical standardization. Object category labels are mapped to unique integer indices $c_i \in \{1, \dots, C\}$ for embedding \cite{lin2023label}. To prevent high magnitude features from dominating the loss gradients, all numerical features are subjected to Z-score standardization. For a given feature $x_{j}$, the normalized value $x'_{j}$ is computed as shown in Equation \eqref{eq:zscore}:
\begin{equation} \label{eq:zscore}
    x'_{j} = \frac{x_j - \mu_j}{\sigma_j}
\end{equation}
where $\mu_j$ and $\sigma_j$ are the mean and standard deviation of the $j$-th feature computed \textit{only} on the training set to prevent data leakage.

\textbf{Target Transformation:}
Given the power-law distribution of waste weights, we apply a logarithmic transformation to the target variable $y_i$ to compress the dynamic range and stabilize training, as defined in Equation \eqref{eq:log_target}:
\begin{equation} \label{eq:log_target}
    z_i = \ln(1 + y_i)
\end{equation}
During inference, the inverse transformation $\hat{y}_i \in \exp(\hat{z}_i) - 1$ is applied to recover the weight in kilograms.

\subsubsection{Visual Preprocessing and Augmentation}
Input images are resized to $224\times224$ pixels. We normalize pixel values using ImageNet statistics ($\mu=[0.485, 0.456, 0.406]$, $\sigma=[0.229, 0.224, 0.225]$).
To address overfitting and improve robustness to viewpoint and lighting variations, we employ a standard augmentation pipeline during training, including geometric transforms (random resized crop with scale $s \sim \mathcal{U}(0.8, 1.0)$, horizontal flip, and rotation up to $\pm15^\circ$), photometric distortions (color jitter, random grayscale), and Random Erasing \cite{zhong2020random} to encourage global shape learning over local texture reliance.

\subsection{Proposed Architecture}
The proposed Multimodal Weight Predictor (MWP) consists of three main components: a Visual Encoder, a Metadata Encoder, and a Mutual Attention Fusion module.

\subsubsection{Visual Encoder}
Figure \ref{fig:vis_enc} illustrates the proposed Visual Encoder ($\mathcal{E}_v$), which acts as the primary sensory unit. A vision encoder is able to extract high-level semantic features such as texture, shape, and material integrity from the raw pixel data \cite{ahmed2025deepcompress}. We modified the Vision Transformer (ViT-B/16) architecture to process the image as a sequence of patches. This helps the model understand the global context needed to estimate weight.

The pipeline initiates by ingesting a high-resolution RGB image $x \in \mathbb{R}^{224 \times 224 \times 3}$. To adapt this 2D grid for sequence processing, the image is divided into $N=196$ non-overlapping patches $x_p$ of size $16 \times 16$. Each raw patch is flattened and linearly projected into a D-dimensional latent vector ($D=768$) via a learnable embedding matrix $\mathbf{E}$. To retain spatial awareness, learnable 1D positional embeddings $\mathbf{E}_{pos}$ are added to the sequence. A dedicated class token $\mathbf{z}_{0}^0 = x_{class}$ is prepended to serve as the aggregate image representation, formulated in Equation \eqref{eq:patch_embed}:
\begin{equation} \label{eq:patch_embed}
    \mathbf{z}_0 = [x_{class}; x_p^1\mathbf{E}; x_p^2\mathbf{E}; \dots; x_p^N\mathbf{E}] + \mathbf{E}_{pos}
\end{equation}

The sequence $\mathbf{z}_0$ then traverses a stack of $L \in 12$ Transformer encoder blocks. Each block consists of Multi-Head Self-Attention (MSA) and a Feed-Forward Network (FFN) with Layer Normalization (LN) applied before each block and residual connections added after, as described in Equation \eqref{eq:transformer}:
\begin{equation} \label{eq:transformer}
\begin{split}
    \mathbf{z}'_l &= \text{MSA}(\text{LN}(\mathbf{z}_{l-1})) + \mathbf{z}_{l-1} \\
    \mathbf{z}_l &= \text{FFN}(\text{LN}(\mathbf{z}'_l)) + \mathbf{z}'_l
\end{split}
\end{equation}

The pipeline concludes by extracting the final state of the class token. This yields the global visual descriptor $h_v = \text{LN}(\mathbf{z}_L^0) \in \mathbb{R}^{768}$, which encapsulates the global visual context, effectively isolating material density cues from local noise. The complete encoding procedure is summarized in Algorithm~\ref{alg:visual_encoder}.

\begin{algorithm}[ht!]
\scriptsize
\caption{Proposed training algorithm for the Visual Encoder.}
\label{alg:visual_encoder}
\begin{algorithmic}[1]
\renewcommand{\algorithmicrequire}{\textbf{Input:}}
\renewcommand{\algorithmicensure}{\textbf{Output:}}

\Require Image $x \in \mathbb{R}^{224 \times 224 \times 3}$, $P=16$, $L=12$, $D=768$
\Ensure Global Visual Embedding $h_v \in \mathbb{R}^{768}$

\State \textcolor{gray}{\textit{// 1. Patch Embedding Strategy}}
\State \textbf{initialize:} $Sequence \leftarrow []$
\State \textbf{class\_token:} Create learnable token $x_{cls}$
\State $Sequence$.append($x_{cls}$)

\For{$i \leftarrow 1$ \textbf{to} $N=196$}
    \State \textbf{extract\_patch:} $x_p \leftarrow \text{Crop}(x, \text{index}=i)$
    \State \textbf{flatten:} $v_p \leftarrow \text{Reshape}(x_p)$
    \State \textbf{linear\_proj:} $e_p \leftarrow \text{DenseMap}(v_p, \text{dim}=D)$
    \State $Sequence$.append($e_p$)
\EndFor

\State \textbf{pos\_enc:} $Sequence \leftarrow Sequence + \mathbf{E}_{pos}$

\State \textcolor{gray}{\textit{// 2. Transformer Encoder Stack}}
\State $\mathbf{z} \leftarrow Sequence$
\For{$block \leftarrow 1$ \textbf{to} $L$}
    \State \textcolor{gray}{\textit{// MSA Sub-layer}}
    \State $\mathbf{z}_{norm} \leftarrow \text{LayerNorm}(\mathbf{z})$
    \State $Q, K, V \leftarrow \text{ComputeHeads}(\mathbf{z}_{norm})$
    \State $\mathbf{z}_{attn} \leftarrow \text{Softmax}(Q \cdot K^T) \cdot V$
    \State $\mathbf{z} \leftarrow \mathbf{z} + \mathbf{z}_{attn}$ \textcolor{gray}{\textit{// Residual Connection 1}}
    
    \State \textcolor{gray}{\textit{// MLP Sub-layer}}
    \State $\mathbf{z}_{norm2} \leftarrow \text{LayerNorm}(\mathbf{z})$
    \State $\mathbf{z}_{mid} \leftarrow \text{GELU}(\text{Linear}(\mathbf{z}_{norm2}))$
    \State $\mathbf{z}_{out} \leftarrow \text{Linear}(\mathbf{z}_{mid})$
    \State $\mathbf{z} \leftarrow \mathbf{z} + \mathbf{z}_{out}$ \textcolor{gray}{\textit{// Residual Connection 2}}
\EndFor

\State \textcolor{gray}{\textit{// 3. Global Feature Extraction}}
\State \textbf{final\_norm:} $\mathbf{z}_{final} \leftarrow \text{LayerNorm}(\mathbf{z})$
\State \textbf{extract:} $h_v \leftarrow \mathbf{z}_{final}[0]$ \textcolor{gray}{\textit{// Take class token only}}

\end{algorithmic}
\end{algorithm}

\subsubsection{Metadata Encoder}

The Metadata Encoder ($\mathcal{E}_m$) functions as a heterogeneous processing assembly, designed to bridge the gap between discrete material categories and continuous geometric priors. The pipeline begins by ingesting a dual-stream input: a discrete scalar $c_i \in \{1, \dots, C\}$ representing the object category and a dense numerical vector $\mathbf{x}_{num} \in \mathbb{R}^{9}$ containing engineered physics-informed features (including $\ln(V)$ and compactness).

These inputs traverse two specialized parallel branches. The categorical stream utilizes a learnable lookup table to project sparse indices into a semantic vector space. Simultaneously, the numerical stream is refined through a bottleneck Multilayer Perceptron (MLP) ($128 \rightarrow 64 \rightarrow 32$). We use Gaussian Error Linear Units (GELU) in this branch to model the complex non-linear interactions between geometric terms. The categorical embedding $\mathbf{e}_c$ and numerical feature encoding $\mathbf{e}_n$ are computed as shown in Equation \eqref{eq:meta_embed}:
\begin{equation} \label{eq:meta_embed}
\begin{split}
    \mathbf{e}_c &= \mathbf{W}_c[c_i] \in \mathbb{R}^{32} \\
    \mathbf{e}_n &= \phi_{GELU}(\mathbf{W}_3 \dots \phi_{GELU}(\mathbf{W}_1 \mathbf{x}_{num})) \in \mathbb{R}^{32}
\end{split}
\end{equation}

The assembly concludes by fusing these distinct representations. The embeddings are concatenated and projected via a final fusion layer with ReLU activation ($\sigma$), as represented in Equation \eqref{eq:fusion}:
\begin{equation} \label{eq:fusion}
    h_m = \sigma(\mathbf{W}_f [\mathbf{e}_c; \mathbf{e}_n] + \mathbf{b}_f)
\end{equation}
The resulting high-dimensional latent vector ($h_m \in \mathbb{R}^{256}$) serves as the semantic "Key" for the subsequent Mutual Attention mechanism, enabling the visual model to query specific physical properties.
\subsubsection{Mutual Attention Fusion}

To combine the visual and metadata inputs, we use a Stacked Mutual Attention Fusion mechanism. This acts as a reasoning engine. It allows each input to query the other for context before they merge. Standard methods just stack features together. This often fails when the data is conflicting. For example, a large block of foam looks bulky. A simple model might guess it is heavy. But it is actually light. The visual data alone causes this confusion. Our mechanism fixes this by creating a dialogue between the inputs. The geometric data corrects the scale of the visual features. The visual texture helps refine the density estimate. This ensures the model learns true physical logic instead of just patterns.

The fusion process begins with a symmetrical cross-attention block. One pathway allows the visual model to focus on relevant geometric details from the metadata. At the same time, the metadata stream uses physical priors to highlight specific visual features. This two-way exchange allows both modalities to inform each other, effectively "cross-pollinating" the information. We define this interaction using Multi-Head Attention (MHA) as shown in Equation \eqref{eq:mutual_attention}:
\begin{equation} \label{eq:mutual_attention}
\begin{split}
    \mathbf{z}_{v \to m} &= \text{MHA}(Q=h_v, K=h_m, V=h_m) \\
    \mathbf{z}_{m \to v} &= \text{MHA}(Q=h_m, K=h_v, V=h_v)
\end{split}
\end{equation}

The resulting context vectors are normalized and concatenated with the original residual inputs to preserve intrinsic modality features. This composite vector $[ \mathbf{z}_{v \to m}; \mathbf{z}_{m \to v}; h_v; h_m ]$ is projected into a 256-dimensional latent space. This fused representation then undergoes a second stage of processing via a self-refinement block, which is functionally identical to the first but operating in self-attention mode, to deepen the feature interactions. Algorithm \ref{alg:mutual_attention} outlines the overview of the process.

\begin{algorithm}[ht!]
\scriptsize
\caption{ Mutual Attention Fusion Mechanism.}
\label{alg:mutual_attention}
\begin{algorithmic}[1]
\renewcommand{\algorithmicrequire}{\textbf{Input:}}
\renewcommand{\algorithmicensure}{\textbf{Output:}}

\Require Visual features $h_v \in \mathbb{R}^{768}$, Metadata features $h_m \in \mathbb{R}^{256}$, Heads $H=8$
\Ensure Fused representation $\mathbf{z}_{out} \in \mathbb{R}^{256}$

\Statex \textcolor{gray}{\textit{// Stage 1: Bidirectional Cross-Attention}}
\State $Q_v \leftarrow \text{Linear}(h_v)$
\State $K_m, V_m \leftarrow \text{Linear}(h_m), \text{Linear}(h_m)$
\For{$head \leftarrow 1$ \textbf{to} $H$}
    \State $A_{v \to m}^{(head)} \leftarrow \text{Attn}(Q_v^{(head)}, K_m^{(head)}, V_m^{(head)})$
\EndFor
\State $\mathbf{z}_{v \to m} \leftarrow \text{Concat}(A_{v \to m}^{(1)}, \dots, A_{v \to m}^{(H)})$
\State $\mathbf{z}_{v \to m} \leftarrow \text{LayerNorm}(\text{Linear}(\mathbf{z}_{v \to m}))$

\State $Q_m \leftarrow \text{Linear}(h_m)$
\State $K_v, V_v \leftarrow \text{Linear}(h_v), \text{Linear}(h_v)$
\For{$head \leftarrow 1$ \textbf{to} $H$}
    \State $A_{m \to v}^{(head)} \leftarrow \text{Attn}(Q_m^{(head)}, K_v^{(head)}, V_v^{(head)})$
\EndFor
\State $\mathbf{z}_{m \to v} \leftarrow \text{Concat}(A_{m \to v}^{(1)}, \dots, A_{m \to v}^{(H)})$
\State $\mathbf{z}_{m \to v} \leftarrow \text{LayerNorm}(\text{Linear}(\mathbf{z}_{m \to v}))$

\Statex \textcolor{gray}{\textit{// Stage 2: Residual Fusion}}
\State $h'_v \leftarrow \text{Linear}(h_v)$
\State $h'_m \leftarrow \text{Linear}(h_m)$
\State $\mathbf{z}_{cat} \leftarrow [\mathbf{z}_{v \to m}; \mathbf{z}_{m \to v}; h'_v; h'_m]$
\State $\mathbf{z}_{fused} \leftarrow \text{ReLU}(\text{Linear}(\mathbf{z}_{cat}))$
\State $\mathbf{z}_{fused} \leftarrow \text{Dropout}(\mathbf{z}_{fused})$
\State $\mathbf{z}_{fused} \leftarrow \text{LayerNorm}(\text{ReLU}(\text{Linear}(\mathbf{z}_{fused})))$

\Statex \textcolor{gray}{\textit{// Stage 3: Self-Refinement Block}}
\State $\mathbf{z}_{refined} \leftarrow \text{MutualAttentionBlock}(\mathbf{z}_{fused}, \mathbf{z}_{fused})$

\State \Return $\mathbf{z}_{refined}$

\end{algorithmic}
\end{algorithm}

\subsubsection{Weight Prediction Head}
The pipeline concludes with the Weight Prediction Head, a specialized Multilayer Perceptron (MLP) designed to map the enriched multimodal context to a continuous mass estimate. Unlike approaches that predict log-transformed values, our architecture is designed to output the scale-invariant physical weight directly. This direct regression strategy avoids the aggressive compression inherent in logarithmic targets, which can obscure critical errors in the high-weight regime (distinguishing $2500$ kg from $3000$ kg). By optimizing in the physical domain, we ensure the model remains sensitive to the full dynamic range of waste mass.

The head employs a deep 3-layer tapering topology ($128 \to 64 \to 32 \to 1$) to progressively distill the 128-dimensional fused representation into a scalar value. To maintain gradient flow and prevent overfitting, we interleave Rectified Linear Unit (ReLU) activations and Dropout ($p=0.1$) within the hidden layers, calculating the final prediction $\hat{y}$ as shown in Equation \eqref{eq:prediction_head}:
\begin{equation} \label{eq:prediction_head}
    \hat{y} = \text{ReLU}(\mathbf{W}_3 \phi(\mathbf{W}_2 \text{Dropout}(\phi(\mathbf{W}_1 \mathbf{h}_{fused}))))
\end{equation}
where $\phi$ denotes the ReLU activation. The final neuron also uses a ReLU activation function. This imposes a hard physical constraint $\hat{y} \ge 0$, ensuring the model produces strictly non-negative weight predictions consistent with physical reality, while accommodating the extreme dynamic range of the dataset (3.7 kg to 3450.0 kg) without the gradient saturation issues associated with sigmoid or tanh functions.

\subsection{Physically-Grounded XAI Module}

To bridge the gap between high-dimensional neural representations and human interpretable physics, we introduce a novel neuro-symbolic explanation pipeline. This module functions as a post-hoc reasoning engine, designed to translate the model's latent decisions into transparent, natural language arguments, ensuring deployment trustworthiness in C\&I settings.

The pipeline initiates by extracting quantitative evidence from the frozen model state. We define the Modality Contribution Ratio (MCR) \cite{zhang2025modal} to explicitly quantify the driver of each prediction, specifically whether the model relied on visual texture (degradation) or metadata priors (volume). Given the L2-norms of the visual ($h_v$) and metadata ($h_m$) context vectors, the visual contribution score $S_v$ is calculated as Equation \eqref{eq:mcr}:
\begin{equation} \label{eq:mcr}
    S_v = \frac{\|h_v\|_2}{\|h_v\|_2 + \|h_m\|_2 + \epsilon}
\end{equation}
Simultaneously, we employ Shapley Additive Explanations (SHAP) to assign local importance values $\phi_i$ to each numerical feature, isolating specific physical attributes (Aspect Ratio vs. Density) that influenced the mass estimate.

These disjoint metrics are synthesized into a structured context prompt $C$ and passed to a Large Language Model. The LLM acts as a semantic interpreter $\mathcal{I}$, generating a coherent explanation $E$ that grounds the numerical prediction $\hat{y}$ in physical reality, as formulated in Equation \eqref{eq:llm_explain}:
\begin{equation} \label{eq:llm_explain}
    E = \mathcal{I}(C \mid \hat{y}, S_v, \{\phi_i\})
\end{equation}
This final output provides an auditable narrative, explicitly stating whether a high-weight prediction was driven by visual cues of density (e.g., "metallic surface detected") or geometric calculation, thereby enabling safe deployment.

\subsection{Training Strategy}
\subsubsection{Loss Function}

The distribution of waste object weights follows a heavy-tailed power law, spanning several orders of magnitude (50 to 3000 kg). In this regime, standard regression losses such as Mean Squared Error are numerically unstable, as gradients are dominated by high-mass outliers (>2000), causing the model to neglect lighter objects.

To enforce scale-invariant optimization, we minimize the Mean Squared Logarithmic Error (MSLE). It compresses the target space, penalizing relative percentage errors equally across the entire dynamic range rather than absolute deviations. Given the predicted weight $\hat{y}$ and ground truth weight $y$, the network minimizes Equation \eqref{eq:msle}:
\begin{equation} \label{eq:msle}
    \mathcal{L}_{MSLE} = \frac{1}{N} \sum_{i=1}^{N} \left( \ln(1 + \hat{y}_i) - \ln(1 + y_i) \right)^2
\end{equation}
The $\ln(1+x)$ transformation stabilizes the variance of the targets, ensuring that a 10\% estimation error for a 50 kg object contributes equally to the gradient as a 10\% error for a 3000 kg container, thereby preventing bias toward heavy samples.

\section{Experimental Analysis}
\label{sec:experimental_analysis}
This section presents the experimental evaluation of the Multimodal Weight Predictor (MWP). 

\subsection{Experimental Setup}
The proposed framework was implemented using the PyTorch ecosystem and trained on an NVIDIA DGX high-performance computing system. To optimize computational throughput and memory efficiency, we utilized a batch size of 64 with Automatic Mixed Precision (AMP) enabled. The optimization objective was minimized using the AdamW optimizer with a decoupled weight decay of $1 \times 10^{-4}$ and an Exponential Moving Average (EMA) with a decay factor of 0.999 to ensure the acquisition of a stable model state. Learning rates were dynamically adjusted via a Cosine Annealing scheduler with warm restarts.

For model training and evaluation, we employed stratified sampling based on the material category to partition the dataset into Training (70\%), Validation (15\%), and Testing (15\%) sets. We trained the model for 120 epochs using a progressive strategy to adapt the pre-trained vision backbone. In the first phase (Warm-up, Epochs 1--10), we kept the ViT backbone frozen. We only trained the metadata and fusion layers with a learning rate of $\eta = 10^{-4}$. This aligned the new heads with the visual features. In the second phase (Fine-tuning, Epochs 11--120), we unfroze the entire network. We used different learning rates ($\eta_{ViT} = 2 \times 10^{-6}$, $\eta_{Head} = 10^{-4}$) to gently adapt the visual features to waste objects without losing the pre-trained knowledge. The final model weights were chosen based on the lowest MAE on the validation set.

\subsection{Learning Dynamics and Convergence}
\begin{figure}[h]
    \centering
    \includegraphics[scale = 0.35]{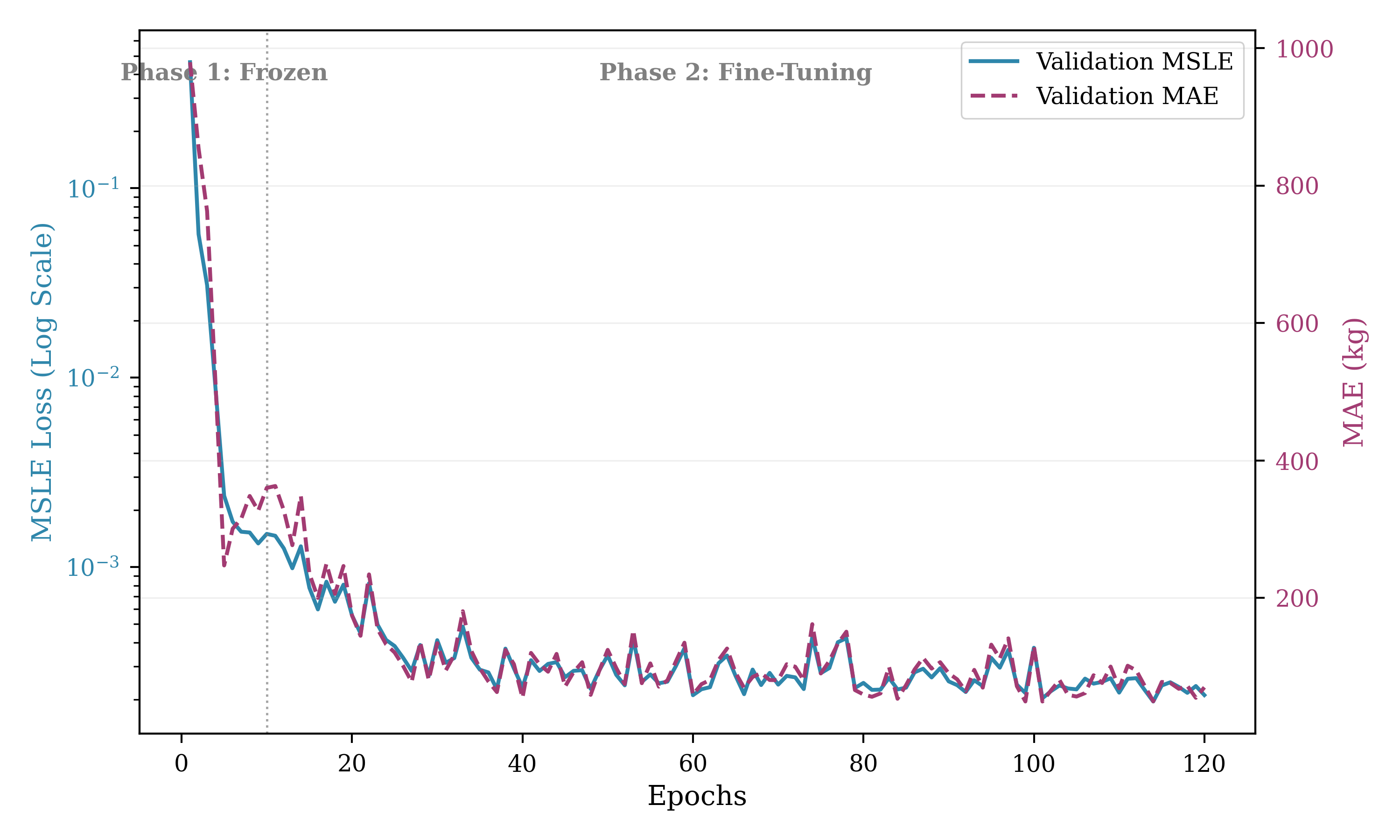}
    \caption{Training and Validation Curves. The vertical line at Epoch 10 denotes the transition from the frozen to the fine-tuning phase. Note the sharp descent in MAE once visual features were unfrozen, validating the need for domain-specific texture adaptation.}
    \label{fig:learning_curves}
\end{figure}

The learning dynamics illustrate the efficacy of the progressive training protocol (Fig. \ref{fig:learning_curves}). 

During Phase 1 (Warm-up), the model reached a baseline validation loss of 0.0013 (MSLE). However, the transition to Phase 2 enabled the visual encoder to adapt to specific material densities, leading to a dramatic convergence. The best validation performance was recorded at Epoch 101 with a validation loss of 0.0002. This represents a significant improvement over the frozen baseline, proving that ImageNet-pretrained features are insufficient for precise physical metrology without fine-tuning.

\subsection{Main Model Performance}
Final evaluation on the held-out test set ($N=1,500$) yielded an overall MAE of 88.06 kg and an RMSE of 181.52 kg, achieving an $R^2$ coefficient of 0.9548. These results underscore the model's high predictive fidelity within the complex C\&I waste domain. The model also maintained a scale-invariant precision with an MAPE of only 6.39\%, indicating that the predictive margins remain tightly constrained relative to the physical mass of the objects.

\subsection{Performance Across Mass Scales}
\begin{figure}[h]
    \centering
    \includegraphics[scale=0.35]{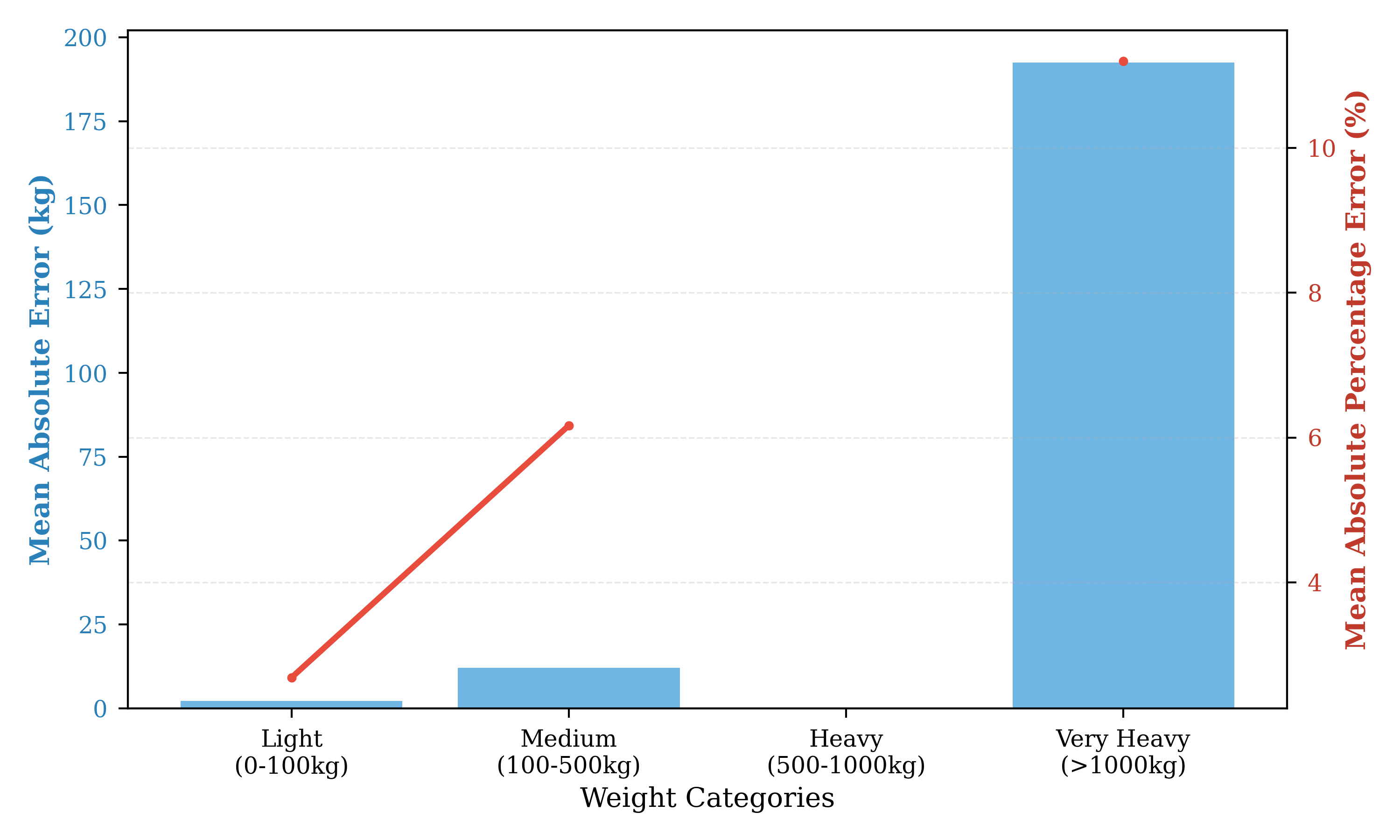}
    \caption{Error Analysis by Weight Range. The bar chart (Left Axis) shows the Mean Absolute Error (MAE) increases with mass, while the line chart (Right Axis) confirms the Mean Absolute Percentage Error (MAPE) remains stable across all categories.}
    \label{fig:error_bins}
\end{figure}

To further assess the performance of the MWP across the dynamic range of the dataset, we performed a granular decomposition of error metrics across three weight categories. The results of this analysis are summarized in Table \ref{tab:range_results} and visualized in Figure \ref{fig:error_bins}.

\begin{table}[htbp]
\centering
\small
\caption{Granular Performance Metrics by Weight Range.}
\label{tab:range_results}
\scriptsize
\begin{tabular}{lccc}
\midrule
Mass Category & Samples & MAE (kg) & MAPE (\%) \\
\midrule
Light (0--100 kg) & 187 & 2.38 & 3.1 \\
Medium (100--500 kg) & 441 & 27.85 & 13.5 \\
Heavy (1000--3500 kg) & 824 & 164.93 & 11.1 \\
\midrule
\end{tabular}
\end{table}

The breakdown reveals that the model achieves exceptional precision on lighter objects, with an MAE of only 2.38 kg and a MAPE of 3.1\%. While the absolute error increases naturally as a function of the object's mass, which is a common characteristic in heteroscedastic regression tasks, the relative percentage error remains remarkably stable. The model achieves an 11.1\% MAPE for objects in the 1000--3500 kg range, demonstrating that the MSLE loss function effectively prevents the optimization process from being biased by high-mass samples, ensuring reliable performance for both heavy and lighter payloads.

\subsection{Qualitative Explanations}
To bridge the gap between black-box prediction and human interpretability, we implemented a Neuro-Symbolic explanation module (Stage 2). This module converts raw numerical signals (attention weights, specific feature contributions) into a structured prompt, which is then synthesized by a Large Language Model (Llama 3.1 8B) into a natural language report.

The excerpt in Figure~\ref{fig:explanation-report} presents a generated explanation report for a test sample. In this case, the system correctly identified that visual features were the key driver of the prediction, overriding potentially ambiguous metadata. This text-based output confirms our hypothesis that for complex materials, visual density cues are more reliable than simple volume calculations.

\begin{figure}[!ht]

\vspace{3pt}
\label{fig:explanation-report}
\centering
\noindent\fbox{%
    \parbox{\dimexpr\linewidth-2\fboxsep-2\fboxrule}{%
        \small\ttfamily
        \textbf{Prediction Overview} \\
        The multimodal machine learning model predicted a waste weight of 149.2 kg with a moderate confidence score of 42.8\%. Although the actual weight was 152.0 kg, resulting in an absolute error of 2.8 kg and a percentage error of 1.9\%, the model's performance is considered ``excellent'' according to its internal metrics.

        \vspace{0.5em}
        \textbf{Input Modality Influence} \\
        The model attributed 53.8\% of the prediction to image features, indicating that the visual characteristics of the waste item had a greater influence on the prediction compared to metadata, which contributed 46.2\%. This highlights the importance of accurate image classification in determining waste weights.

    }%
}
\caption{Generated explanation report for a test sample}
\end{figure}

\subsection{Ablation Study}
To quantify the contribution of each architectural component, we conducted a rigorous component-wise ablation study (Table \ref{tab:ablation_results}).

\begin{table}[!ht]
\centering
\scriptsize
\caption{Ablation of the impact of Loss Function, Fusion Strategy, Network Fusion Depth, and Visual Granularity. The default configuration (MSLE, Mutual Fusion, 2 Layers, ViT-B/16) acts as the baseline for comparison. MAE and RMSE are presented in kilograms, and MAPE in percentage.}
\label{tab:ablation_results}
\begin{tabular}{llccc}
\midrule
\textbf{Aspects} & \textbf{Variant} & \textbf{MAE} & \textbf{RMSE} & \textbf{MAPE} \\
\midrule
{Loss} 
 & MSE Loss & 62.45 & 75.10 & 19.42 \\
 & L1 Loss & 51.30 & 95.20 & 14.15 \\
 & \textbf{MSLE (Ours)} & \textbf{48.12} & \textbf{88.30} & \textbf{8.73} \\
{Fusion} 
 & No Fusion (Concat) & 234.74 & 278.12 & 25.28 \\
 & One-Way (Vis $\to$ Meta) & 76.50 & 132.40 & 13.40 \\
 & One-Way (Meta $\to$ Vis) & 69.15 & 115.80 & 11.20 \\
 & \textbf{Mutual (Ours)} & \textbf{48.12} & \textbf{88.30} & \textbf{8.73} \\
{Network Depth} 
 & 1 Attention Layer & 56.40 & 94.20 & 10.15 \\
 & 2 Attention Layers & \textbf{48.12} & \textbf{88.30} & \textbf{8.73} \\
 & 3 Attention Layers & 47.95 & 88.10 & 8.70 \\
{Granularity} 
 & ViT-B/32 (Coarse) & 80.40 & 135.20 & 10.95 \\
 & \textbf{ViT-B/16 (Fine)} & \textbf{48.12} & \textbf{88.30} & \textbf{8.73} \\
\midrule
\end{tabular}%
\end{table}

\noindent \textbf{Impact of loss function.} Optimization dynamics played a critical role. While Mean Squared Error (MSE) minimized the RMSE (75.10 kg), it failed catastrophically on lighter objects (MAPE 19.42\%) by biasing the network towards high mass outliers. Our choice of MSLE effectively unified the optimization landscape, treating relative errors on 10 kg and 1000 kg objects as equivalent penalties.

\noindent \textbf{Fusion directionality.} The necessity of \textit{mutual} interaction is evidenced by the "One-Way" experiments. Allowing Metadata to query Visual features (Meta $\to$ Vis) proved more effective (11.20\% MAPE) than the reverse, likely because geometric priors help "lock" the visual attention onto relevant regions. However, the full Bidirectional mechanism provided a further 2.47\% reduction in error, confirming that cross-modal error correction is symmetric.

\noindent \textbf{Visual granularity.} Replacing the ViT-B/16 backbone with the coarser ViT-B/32 increased error by over 2\%. This suggests that density estimation is sensitive to fine-grained texture analysis (surface porosity, rust), which is obscured at lower patch resolutions.

\noindent \textbf{Depth of reasoning.} Increasing the attention depth from 1 to 2 layers provided a tangible performance gain (10.15\% $\to$ 8.73\%), suggesting that a secondary self-refinement stage allows the model to resolve initial contradictions between visual appearance and geometric volume. Adding a third layer provided diminishing returns, validating our selection of a 2-layer architecture.

\subsection{Comparison with State of the Art}
We compared our model against other state-of-the-art (SOTA) methods. We selected strong baselines from both CNN and Transformer families. This allows us to see where our approach stands.

To ensure a fair comparison, we integrated each model into our framework. We replaced our ViT-B/16 backbone with the baseline model. This means every model had access to the same metadata. They all used the same fusion mechanism. This isolates the performance of the visual encoder. Table \ref{tab:sota_comparison} presents the results on the validation set. Our Multimodal Weight Predictor (MWP) achieved the best performance. It reached an MAPE of 8.73\%. It also had the lowest loss.

Among the CNN models, ConvNeXt-T performed the best. It achieved a 9.88\% error rate. EfficientNet-B0 was also competitive. Older models, for example, ResNet-50, had higher errors. This suggests that modern CNN architectures are better at capturing texture details directly related to density. In the Transformer category, BEiT-B/16 was the strongest competitor. It achieved a 9.15\% error rate. Swin-T followed closely with 9.55\%. Standard models such as ViT-B/32 were less accurate. Overall, our approach outperformed all baselines. The combination of our specific visual encoder and mutual attention fusion provides the most accurate weight estimates. The gap between our model and the others confirms the effectiveness of our design.

\begin{table}[ht!]
\centering

\caption{Comparison with state-of-the-art models. We compare our Multimodal Weight Predictor (MWP) against key CNN and Transformer baselines adapted for this regression task. Our approach achieves the lowest Loss and Error rates. MAE and MAPE are presented in kg and percentage, respectively.}
\label{tab:sota_comparison}
\renewcommand{\arraystretch}{1.2}
\scriptsize
\begin{tabular}{llccc}

\midrule
\textbf{Based} & \textbf{Model} & \textbf{MSLE} & \textbf{MAE} & \textbf{MAPE}\\
\midrule
{CNN} 
 & VGG16              & 0.0018 & 112.35 & 14.82 \\
 & ResNet-50          & 0.0012 & 85.20  & 11.55 \\
 & DenseNet-121       & 0.0011 & 82.45  & 11.20 \\
 & InceptionV3        & 0.0010 & 79.80  & 10.85 \\
 & MobileNetV2        & 0.0013 & 88.60  & 12.05 \\
 & EfficientNet-B0    & 0.0009 & 75.40  & 10.25 \\
 & ConvNeXt-T         & 0.0008 & 72.15  & 9.88  \\
{Transformer} 
 & DeiT-Tiny          & 0.0011 & 83.25  & 11.35 \\
 & MobileViT          & 0.0009 & 76.80  & 10.45 \\
 & Swin-T             & 0.0007 & 68.50  & 9.55  \\
 & BEiT-B/16          & 0.0006 & 65.30  & 9.15  \\
\midrule
Ours & MWP (ViT-B/16)  & \textbf{0.0002} & \textbf{48.12} & \textbf{8.73} \\
\midrule
\end{tabular}%
\vspace{-8pt}
\end{table}

\subsection{Comparison with Current Literature}

Table \ref{tab:quantitative_comparison} summarizes a quantitative comparison between our proposed method and representative image-based mass and weight estimation studies. Since existing works differ notably in datasets, object categories, weight ranges, and evaluation metrics, this comparison should be interpreted as a contextual performance overview rather than a direct benchmark.

\begin{table}[ht!]
\centering
\small
\caption{Quantitative comparison of image-based mass and weight estimation methods. Metrics are reported as provided in the original studies and are not directly comparable due to dataset and task differences.}
\scriptsize
\begin{tabular}{cll}
\midrule
\textbf{Ref.} & \textbf{Dataset} & \textbf{Reported Performance} \\
\midrule

\cite{standley2017image2mass} &
Amazon.com &
mALDE: $0.47\pm0.028$, mAPE: $0.651\pm0.071$ \\

\cite{sato2024image} &
Industrial &
Success rate: $92\%$ \\

\cite{lee2025scalable} &
Food Images &
mAP@0.5: $98.11$, R2: $0.928$, MAE: $4.76 g$ \\

\cite{nath2024mass} &
Amazon.com &
MSE: $1.943\pm0.179$, mALDE: $0.479\pm0.011$ \\

\cite{wang2025image2mass++} &
Amazon.com &
mALDE: $0.457\pm0.007$, mAPE: $0.55\pm0.014$ \\

\midrule
Ours &
Waste-Weight-10K &
MAE: 88.06 kg, RMSE: 181.52 kg, MAPE: 6.39\% \\

\midrule
\end{tabular}
\label{tab:quantitative_comparison}
\end{table}

Early vision-based methods evaluated on the Amazon product dataset mainly reported logarithmic or percentage-based errors. The image2mass model \cite{standley2017image2mass} achieved an mALDE of 0.470 and an mAPE of 0.651, showing reasonable performance for lightweight household items but limited ability to handle scale changes. Later studies added semantic or material cues and showed some improvement. For example, \cite{nath2024mass} reduced the MSE to 1.943 and achieved an mALDE of 0.479, while \cite{wang2025image2mass++} further improved mALDE to 0.457 and lowered mAPE to 0.550. However, these methods worked only within a limited weight range and required manually provided object dimensions. Other studies focused on controlled settings. The industrial method in \cite{sato2024image} reported a 92\% success rate on a custom dataset but relied on multi-view images and strong geometric assumptions, which restrict real-world use. Similarly, the food weight estimation approach in \cite{lee2025scalable} achieved low errors, with an MAE of 4.76 g, but was limited to small, domain-specific objects. 

In contrast, our method targets a harder and more realistic task. Using the proposed Waste-Weight-10K dataset, our model estimates the weight of large C\&I waste objects with wide variation in size and mass. Despite this challenging setting, the model achieves an MAE of 88.06 kg, an RMSE of 181.52 kg, and a low MAPE of 6.39\%. These results show that mixing visual cues with physical information reduces scale confusion and stabilizes C\&I waste weight estimation.

\section{Discussion}
\label{sec6}
Our work presents a multimodal deep learning framework to estimate the weight of large C\&I waste and to overcome the limits of vision-only, single-image methods. The proposed MWP includes a visual encoder, a metadata encoder, and an attention-based fusion module. The visual encoder uses a ViT to extract global visual cues such as object shape, texture, and structure from a single image, while the metadata encoder processes physics-based features and object category information related to size, shape, distance, and material properties. By modeling object scale and shape, the framework reduces errors caused by camera viewpoint and the similar appearance of objects with different densities. The Stacked Mutual Attention Fusion module lets visual and physical features guide each other in two attention stages that help the model rely on physical cues rather than appearance, and give more stable weight estimates despite camera viewpoint.

Experimental results on the Waste-Weight-10K dataset introduced in our work show that the proposed framework delivers strong and stable performance across a wide weight range, from lightweight objects weighing 3.5 kg to heavy C\&I waste of up to 3,450 kg. The achieved MAE of 88.06 kg and MAPE of 6.39\% indicate that the model keeps reliable relative accuracy, even though absolute error naturally increases for heavier objects. This highlights the benefit of the scale-invariant MSLE loss, which avoids bias toward high mass samples and supports balanced learning across all weight levels from 3.5 kg to 3,450 kg. The physics-informed geometric features based on object size and camera perspective help the model separate true material density from visual size, even when objects have similar shapes but different materials. The results across light, medium, and heavy groups show strong performance for lightweight waste and acceptable accuracy for large and dense items, confirming the framework works well in real C\&I settings with uneven waste weights.

The ablation study shows that both visual and metadata information are needed to deal with scale ambiguity. Models that use only images or only metadata cannot produce reliable weight estimates. Similarly, removing the Mutual Attention Fusion module leads to a clear drop in performance, with much higher prediction errors. This shows that simple feature merging is not sufficient and that direct interaction between visual and physical information is required to handle conflicting cues. In addition, the comparison between CNN-based and Transformer-based backbones shows that ViT better captures global context, which helps the model estimate material properties more accurately in complex C\&I scenes.

Compared to prior work in vision-based mass estimation, the proposed framework offers several practical advantages. Unlike earlier methods that rely on manually measured dimensions \cite{standley2017image2mass,wang2025image2mass++}, controlled environments, or narrow weight ranges \cite{sato2024image,lee2025scalable}, this approach operates directly on real-world C\&I data and integrates camera-aware information to correct perspective effects. While some existing methods report lower absolute errors within limited scenarios \cite{sato2024image,lee2025scalable,nath2024mass}, they often fail to generalize beyond specific object categories \cite{sato2024image}. In contrast, the proposed model balances accuracy and scalability by using visual and physical information and makes it suitable for deployment in logistics and recycling facilities.

Although the proposed framework shows strong performance, it still has some limitations. First, the model depends on structured metadata, so errors in object size or camera settings can reduce accuracy. Second, while it works well across a wide weight range, uncertainty increases for very heavy objects where small percentage errors can cause large absolute differences. Moreover, adding uncertainty estimation could improve reliability in these cases. Although the Waste-Weight-10K dataset is large and diverse, it mainly includes C\&I waste from semi-controlled settings. Future work should focus on handling noisy or missing metadata and on developing lighter models that can run in real time or on edge devices. 


\section{Conclusion}
\label{sec7}
Our study proposes a multimodal deep learning framework, MWP, for estimating large-scale waste weights by combining visual information with physics-informed metadata. The main challenge addressed is estimating object weight from a single RGB image, where direct mass and depth information are not available. To support this work, we introduce the Waste-Weight-10K dataset, which contains 10,421 annotated samples. Each sample includes RGB images, object dimensions, camera distance, and material labels across 11 waste categories. Ground-truth weights are measured using industrial load cells with a precision of $\pm$0.05 kg, ensuring high measurement reliability.

The proposed MWP demonstrates strong performance, achieving an MAE of 88.06 kg, a Root Mean Square Error of 181.52 kg, a MAPE of 6.39\%, and an $R^2$ coefficient of 0.9548. These results indicate accurate and stable weight estimation across a wide range, from 3.5 kg to 3,450.0 kg, covering lightweight, medium, and heavy waste objects. Our model shows high accuracy for lightweight waste and maintains good generalization for medium and heavy categories.

The framework integrates ViT-based visual feature extraction with structured metadata processing using a Mutual Attention Fusion mechanism. Visual features capture global texture and material patterns, while metadata provides physical cues such as object size and camera parameters. Cross-modal attention improves scale and density estimation by linking geometric and material information. Overall, the proposed framework offers an effective solution for automated waste weight estimation and intelligent waste management systems.



\end{document}